# Deformable Voxel Grids for Shape Comparisons

Raphaël Groscot, Laurent D. Cohen
University Paris-Dauphine, PSL Research University, CEREMADE
CNRS, UMR 7534, 75016 Paris, France

## ABSTRACT

We present Deformable Voxel Grids (DVGs) for 3D shapes comparison and processing. It consists of a voxel grid which is deformed to approximate the silhouette of a shape, *via* energy-minimization. By interpreting the DVG as a local coordinates system, it provides a better embedding space than a regular voxel grid, since it is adapted to the geometry of the shape. It also allows to deform the shape by moving the control points of the DVG, in a similar manner to the Free Form Deformation, but with easier interpretability of the control points positions. After proposing a computation scheme of the energies compatible with meshes and pointclouds, we demonstrate the use of DVGs in a variety of applications: correspondences via cubification, style transfer, shape retrieval and PCA deformations. The first two require no learning and can be readily run on any shapes in a matter of minutes on modest hardware. As for the last two, they require to first optimize DVGs on a collection of shapes, which amounts to a pre-processing step. Then, determining PCA coordinates is straightforward and brings a few parameters to deform a shape.

## 1. INTRODUCTION

Three dimensional shapes are difficult to compare to one another, because they lack a good, canonical basis of representation. On the one hand are surfacic representations: pointclouds and meshes. Pointclouds are unordered sets, so comparing two requires a registration step which essentially determines point-to-point correspondences. This matching can operate under different assumptions: for instance, ICP[3] and variants[24] for rigid affine transformations, or Earth Mover's Distance[8] for minimal transport-cost one-to-one matching. The same holds for meshes, but the explicit connectivity also allows for spectral methods, whose local descriptors derive from eigenvalues of operators such as the Laplacian[21,26]. On the other hand are volumetric representations: voxel grids and signed distance fields (SDF). Being scalar signals in the unit cube of $\mathbb{R}^3$, they indeed have a canonical embedding space. However, a big part of voxel space is non discriminative: the center (*resp.* the border) of the cube is mostly always occupied (*resp.* empty). To capture rich geometric details, they also require a sufficient grid resolution, leading to high dimensional spaces where distances are not meaningful. As for SDFs, they generally serve as a proxy to either render the shape via ray marching[13] or generate a mesh via marching cubes[18]. This is why, in the context of shape editing, both these representations are mainly useful for boolean operations — intersection and union.

Instead of relying solely on their geometry, shapes can be compared using parameters which translate to meaningful properties, such as length, height, etc. In the modeling phase, this can be achieved by parametric surface or volume elements, such as splines and nurbs. But in the analysis phase, these parameters are generally not accessible. This is where shape priors come to play, either in the form of statistical distribution estimation[9], or with methods relying on deformations from a template[17] or between pairs of shapes[12]. These are controlled by the Free Form Deformation[25] (FFD) which, although having volume-preserving properties, offers unintuitive controls. Another approach comes with neural networks, and more especially generative models such as GANs[10] and VAEs[16]. They offer a so-called *latent space* which serves as a canonical, Euclidean parameter space, with a typical dimensionality lower than that of the geometric shape space, amenable to meaningful distances between points. Several works demonstrated how latent arithmetics allows for similarity clustering[22,23], shape analogies[1] and recombinations[7,11]. However, they rely on heavy computations on large datasets, which not only requires powerful hardware and a long training, but also depends on the reconstruction capacity of the chosen generator architecture.

In this paper, we introduce an alternative geometric representation, which we demonstrate is adapted to shape comparisons with very simple computations: the Deformable Voxel Grid (DVG). A DVG is an energy-driven deformation of the unit cube surrounding the shape. Intuitively, if we only consider its outer surface, it consists in an elastic hull which, like a shrinking balloon, embraces the shape (see Figure 1), thus providing a low-level approximation of a given shape.

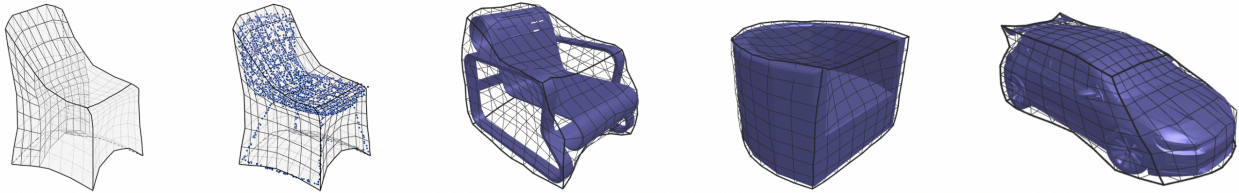

Figure 1. DVGs on different shape categories, for pointclouds and meshes (*left*: volumetric slice showing the inner cells).

It is inspired by the Topological Active Volume[2] (TAV), which is a volumetric extension of active contours[6,15]. The unfamiliar reader can think of active contours as parametric curves which minimize a given energy, typically used for segmenting objects in images[14]. This energy is split into an *intrinsic* term — regularization — and an *extrinsic* term — data fidelity:

$$\mathcal{E}(\theta) = \mathcal{E}_{\text{regularization}}(\theta) + \mathcal{E}_{\text{data}}(\theta) \qquad (1)$$

This energy depends on a set of parameters $\theta$ (the vertices positions for a polygonal curve), thus the minimization problem consists in finding the optimal parameters $\theta^*$. In order to do so, an initial set of parameters $\theta_0$ is determined manually or automatically, and they are iteratively updated via gradient descent. Intuitively, it helps to visualize this dynamics as a curve that evolves through time (each gradient descent step is a time step), hence the name *active* contour. For instance, if the initialization forms a closed loop around the object that needs to be segmented, the intrinsic term corresponds to a contraction force -- along with regularization, and the extrinsic term to an attraction force to the contour of the object. At equilibrium, when the curve matches the contour, these two forces balance each other out. Figure 2 illustrates the evolution of an active contour. The initialization also prevents the active contour from being stuck at an undesired local minimum of the energy, by providing a close enough initial state. An active volume behaves similarly, but in three dimensions. Before drawing the analogy too far, a caveat regarding dimensions has to be noted: whereas in images (2D), an active contour is just a 1-dimensional curve, for 3D shapes, an active volume is 3-dimensional. A 3D analogy of active contours that preserves this dimensionality relationship would be an active *surface*. But Deformable Voxel Grids discretize the embedding space itself, by elastically deforming a regular voxel grid until its outer surface tightly embraces a given shape.

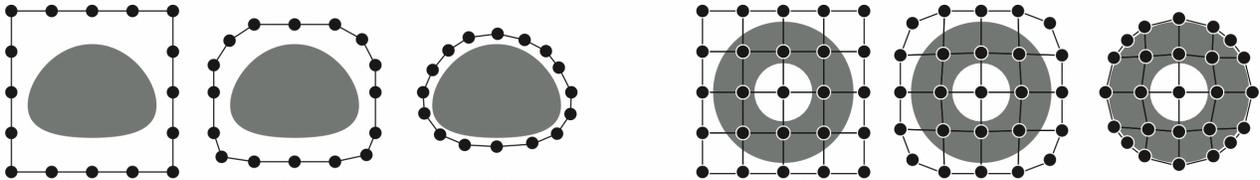

Figure 2. Evolution of active contour (*left*) versus active volume (*right*), depicted in 2D for clarity.

The aim of this paper is not to demonstrate superior performances in a specific application, say correspondences. Rather, it is to present a unified shape representation framework, which offers natural baseline solutions to a surprising variety of problems related to deformation, among which are shape correspondences, but also retrieval of similar shapes, style transfer, and parametric editing (see Section 3). First, we present in Section 2 the DVG model, along with its estimation and post-processing.

## 2. THE DEFORMABLE VOXEL GRID MODEL

A DVG arises from the deformation of the unit cube $\mathbb{R}^3$ which we discretize in a $(r+1) \times (r+1) \times (r+1)$ grid, such that $r$ corresponds to the resolution, i.e. the number of subdivisions along each axis. The vertices of this grid are called *control points*, and their best position is determined by energy-minimization. As a matter of fact, to each state of a DVG corresponds an energy, which measures how well the DVG fits the given shape while remaining smooth. Every control point is connected to its neighbors following the 6-connectivity scheme, and every volume bounded by 8 neighbors is called a *cell*. Even if the faces of a cell are in general non-planar, since 4 points in 3D are not generally coplanar, they are well defined as 3D quadrilaterals. In the end, a DVG forms a hexahedral grid. Its purpose is to offer a consistent representation basis for several shapes belonging to the same class (e.g. chair or car), allowing both:

- **Straightforward comparisons between shapes**: the same cell (identified by its position within the grid) should correspond to the same part of a shape.
- **Easy shape deformation**: after attaching the shape to its optimized DVG, deforming the shape by moving the control points. Because the control points are originally close to the shape, this provides a more intuitive control of the deformation than FFD does.

## 2.1 Grid parameterization and energies

The parameterization of the DVG is taken from the TAV[2]. A continuous DVG is defined as $V(u,v,w) = (x(u,v,w), y(u,v,w), z(u,v,w))$ where $u, v, w \in [0,1]$, and its corresponding energy is then:

$$\mathcal{E}(V) = \int_0^1 \int_0^1 \int_0^1 \mathcal{E}(V(u,v,w)\, du\, dv\, dw \tag{2}$$

where $\mathcal{E}(V(u,v,w))$ decomposes according to (1). Note that our V is discretized in a regular voxel grid, meaning that we only know the value of $V(u,v,w)$ for discrete points in $[0,1]^3$ — corresponding to regular subdivisions of the unit cube — while the rest is interpolated using trilinear interpolation. The DVG is initialized as a regular voxel grid around the shape, and evolves through time, to find a minimal energy state. Equation (1) takes a generic form and must be further detailed. The energy of a given DVG state comprises three different terms with their respective relative weights:

$$\mathcal{E} = \mathcal{E}_{\text{regularization}} + \mathcal{E}_{\text{data}} = \lambda_e \mathcal{E}_{elastic} + \lambda_b \mathcal{E}_{bending} + \lambda_i \mathcal{E}_{inclusion} \tag{3}$$

The regularization energy comprises elastic and bending terms, their minimization tends to respectively shrink the grid and maintain low curvature. The data fidelity term ensures that the outer surface of the DVG is stopped by the shape, by penalizing all points of the shape that lie outside the DVG.

**Elastic and bending energies**. In a similar manner to active contours and topological active volumes, we penalize the squared norm of the first and second derivatives, respectively corresponding to elastic and bending energies:

$$\begin{aligned}\mathcal{E}_{elastic}(V) &= \alpha(|V_u(u,v,w)|^2 + |V_v(u,v,w)|^2 + |V_w(u,v,w)|^2) \\ \mathcal{E}_{bending}(V) &= \beta(|V_{uu}(u,v,w)|^2 + |V_{vv}(u,v,w)|^2 + |V_{ww}(u,v,w)|^2) \\ &\quad + \gamma(|V_{uv}(u,v,w)|^2 + |V_{vw}(u,v,w)|^2 + |V_{uw}(u,v,w)|^2)\end{aligned} \tag{4}$$

where $\alpha, \beta, \gamma$ are coefficients which control the relative importance of each term. The elastic energy induces a contracting force, shrinking the grid inwards, while the bending energy keeps the grid as straight as possible, minimizing its curvature. To compute these first and second derivatives on the discrete parameterization, we use 3D finite differences.

**Inclusion.** The shape should stay inside the DVG: by design, this is the case at initialization, and this energy ensures that it remains the case during the evolution, acting as a barrier against the contraction of the DVG. Note that we only need this barrier on the *surface* of S. In our design and experiments, we only use sparse surface information in the form of a pointcloud uniformly sampled from a triangular mesh (with sampling weights proportional to the triangles' surface area). This allows to use the same DVG model on pointclouds and meshes alike, and simplifies the computation of this energy term. Indeed, to compute $\mathcal{E}_{inclusion}$ we want to determine whether each point sample is located inside V. But this on-or-off signal does not provide gradients for the gradient descent of $\mathcal{E}$, and we would like a stronger penalty for the farthest points. Hence a quantization scheme we devised to efficiently and differentiably estimate this inclusion for all points of S, in parallel using GPU acceleration. To test whether $s \in V$ for any $s \in S$, we resort to a family of balls $\{B_i(p_i, r)\}_{i=1}^l$ of a given radius $r$, located at positions $\{p_i\}_i$, whose union provides an approximation of V, so that:

$$s \in B \Leftrightarrow \exists i, s \in B_i \Leftrightarrow \exists i, |s - p_i|_2 \leq r \Leftrightarrow d(s, \{p_i\}_i) \leq r \Leftrightarrow |s - p^\star| \leq r \tag{5}$$

where $d$ is the distance between a point ($s$) and a set ($\{p_i\}_i, 1 \leq i \leq l$), and $p^\star$ is the $p_i$ closest to $s$. A ball covering of V can easily be obtained by subdividing its cells and taking points at the centers of the newly created subcells (see Figure 3).

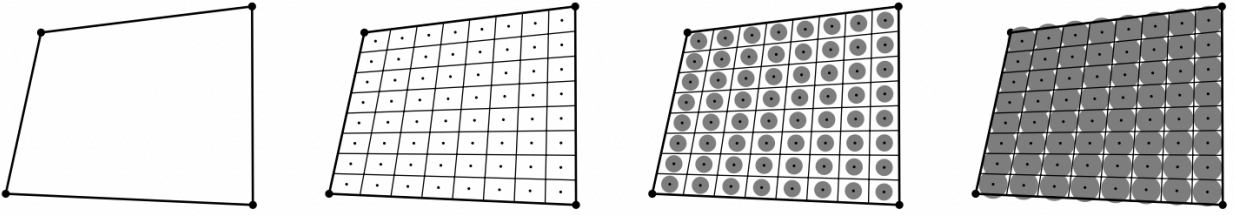

Figure 3. During the computation of the energy $\mathcal{E}_{inclusion}$, each DVG cell is replaced by a set of regularly spaced balls of constant radius. For big enough radii, the union of the balls achieves an approximate covering of the cell.

To have a differentiable loss term, the hard inequality in (5) can be replaced by a soft thresholding thanks to the sigmoid $f_{\theta,\alpha}(x) = \frac{1}{2}\tanh(\alpha(x - \theta)) + \frac{1}{2}$, where $\theta$ is the threshold and $\alpha$ the stiffness (the higher $\alpha$, the more it behaves like a hard thresholding). It follows:

$$\mathcal{L}_\alpha(s \in V) = f_{r,\alpha}\big(d(s, \{p_i\}_i)\big) = f_{r,\alpha}(||s - p^\star||)$$
$$\mathcal{L}_\alpha(S \subset V) = \frac{1}{k}\sum_{i=1}^{k} \mathcal{L}_\alpha(s_k \in D) = \frac{1}{k}\sum_{i=1}^{k} f_{r,\alpha}(||s_k - p_k^\star||) \qquad (6)$$

This equation defines a loss corresponding to $S \subset V$, with the stiffness parameter $\alpha$. This loss is used as $\mathcal{E}_{inclusion}$.

## 2.2 Recursive subdivisions for hierarchical optimization

For a DVG of resolution $r^3$ cells, if r is a power of 2, say $r = 2^p$, we can view the grid as p successive subdivisions of a cube, corresponding to finer and finer resolution levels. This suggests a recursive optimization scheme, starting from a regular cube at resolution level 0. At level k, the position of the control points $v_k$ is optimized via gradient descent on energy $\mathcal{E}$. Then, the grid is subdivided by a factor of 2, which means that each cell is sliced into 8 smaller cells. The created vertices are located at the centers of previous faces and edges, to which a small perturbation (the *residual*) is added with a fading factor $\alpha_k$ (see Figure 4a for an illustration). The recursive relationship between a resolution level $v_k$ and the previous one is then given by:

$$v_k = \text{Sub}(v_{k-1}) + \alpha_k r_k \qquad (7)$$

where $v_k$ corresponds to the vertices of level k, Sub is the subdivision operator, $r_k$ is the residual, and $\{\alpha_k\}_{1 \le k \le p}$ the fading factors. This hierarchical optimization stabilizes the convergence to a solution close to the shape, as our experiments showed (see Figure 4b). To select precision level K for the output, one can set $\{\alpha_k\}_{k \le K}$ to 1 and $\{\alpha_k\}_{k > K}$ to 0. Note that this construction, yielding a collection of $v_k$ by successive super-sampling and perturbations, is similar to a wavelet series decomposition, where each layer encodes details from increasing frequencies (in space) and decreasing amplitudes.

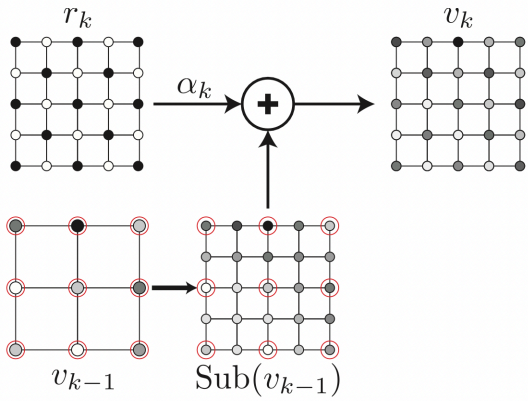
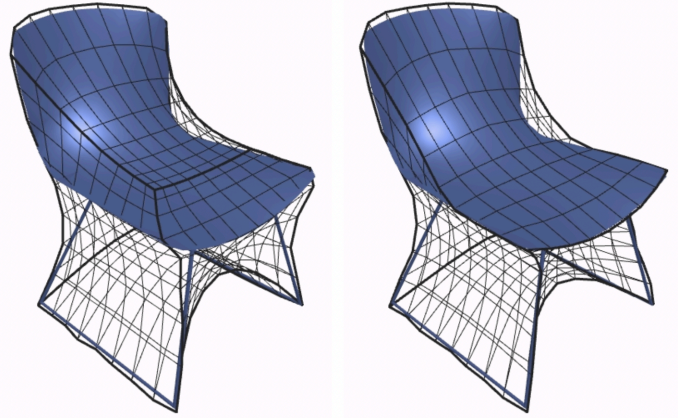

Figure 4. (*Left*) Each level $v_k$ is built as the subdivision of the previous level, added to the next residual, according to Equation (7). The subdivision preserves the key points of a previous level (circled in red) and interpolates the value for the new points. The visualization is simplified by figuring color instead of position (color can be interpreted as the magnitude of a displacement). (*Right*) When the DVG is not optimized hierarchically (left), the edges of the original cube (thicker liners) do not align with the dominant features of the shape (both shown with same number of steps).

### 2.3 Registering a shape to a DVG

Once an optimal DVG V is determined for a shape S, a natural idea is to express S in coordinates $(u, v, w) \in [0,1]^3$ taken from the same parameterization as $V(u, v, w)$. We call this *shape registration*. Given a DVG cell c and a point q inside, we want to determine its *local coordinates*. The trilinear interpolation $f_c$ already used in the hierarchical subdivisions (see Section 2.2) yields a regular mapping from $[0,1]^3$ to the cell:

$$\exists \tilde{u}, \tilde{v}, \tilde{w} \in [0,1]^3 \text{ s.t. } f_c(\tilde{u}, \tilde{v}, \tilde{w}) = q \qquad (8)$$

where $p_1, p_2, ..., p_8$ are the positions of the eight vertices of c. Then, $(\tilde{u}, \tilde{v}, \tilde{w})$ are the local coordinates of point q, so that one would need to invert the trilinear interpolation in order to find them. Then, an affine transformation maps the cell to its correct location within the whole DVG grid system (see Figure 5b).

However, there is no analytical solution to the inversion problem, which requires numerical techniques such as Newton's method. Moreover, this inversion would be local, requiring to solve a different system for each point of the source shape, and would discard any point located outside the DVG. This is why we propose to approximate this inversion using a Thin Plate Spline[4] (TPS), which has the nice property of finding a smooth space deformation based on a set of control points. Here, these will be the control points of the DVG, whose target positions are the equivalent points of a regular voxel grid (see Figure 5b).

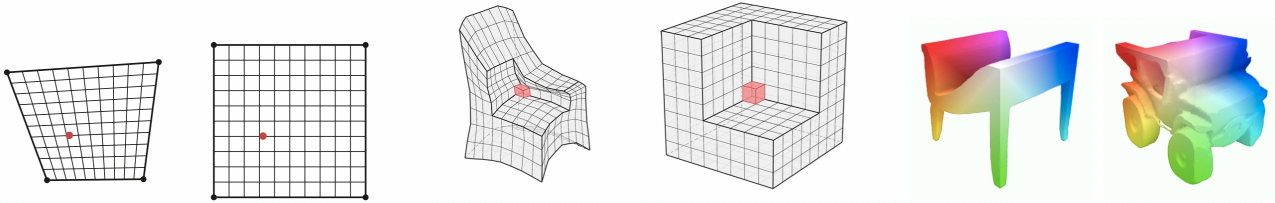

Figure 5. (*Left*) A natural way of setting grid coordinates on a quadrilateral is via bilinear interpolation, which maps regular subdivisons of $[0,1]^2$ onto the quad cell (*left to right*). Determining the local coordinates of a given point within the cell corresponds to inverting this interpolation (*right to left*). (*Center*) The same is done for registering a point inside a DVG, but in 3D, with the inversion of a trilinear interpolation. (*Right*) Examples of cubified shapes (positions in $[0,1]^3$ are color-coded).

## 3. APPLICATIONS AND RESULTS

Our experiments were performed on chair and car models from the ShapeNet[5] repository, with $\lambda_e = 1, \lambda_b = 0.4, \lambda_i = 4, \alpha = \beta = \gamma = 1$.

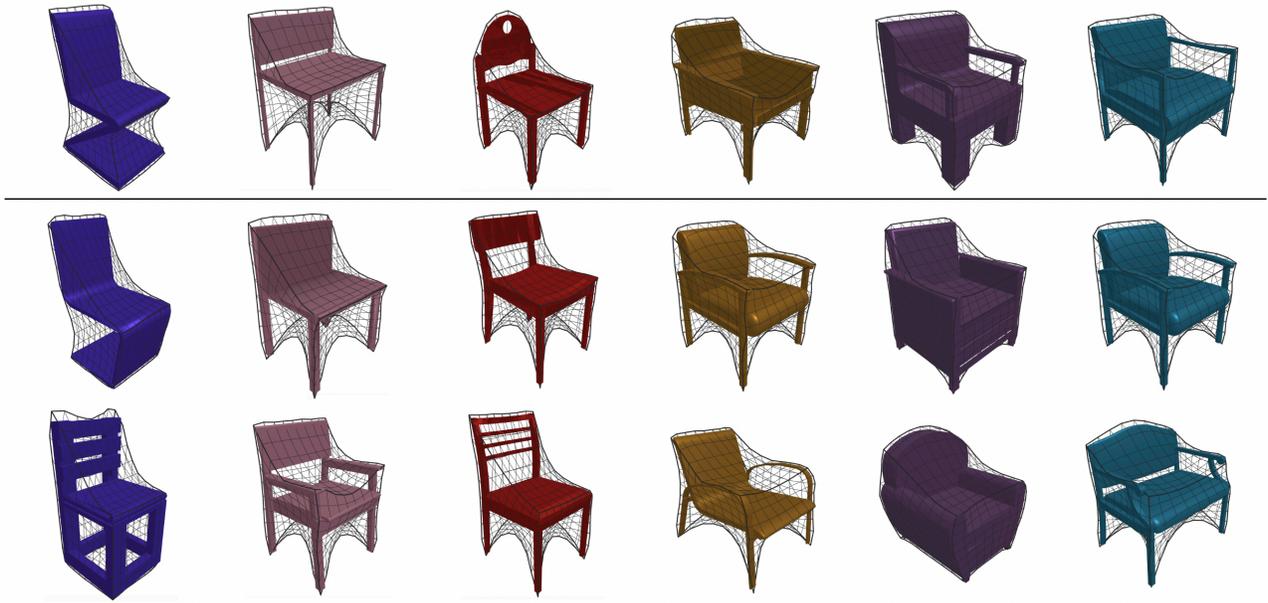

Figure 6. Similarity search via DVG (on 500 models). Top row is the query shape, below are the two nearest neighbors using the outer points of V as a shape descriptor.

### 3.1 Shape descriptor for similarity search

The positions V of a DVG encode a rough global appearance, and can serve as a shape descriptor. A nearest-neighbors search using these descriptors and a simple Euclidean distance, confirms their usefulness in querying a dataset for similar models, as depicted in Figure 6. Observe how this descriptor is blind to differences in topology, but captures the general orientation of semantical parts (back, seat, etc.).

*Shape deformations*: By deforming the DVG of a shape into a different configuration, one can deform the shape itself, either by using the aforementioned registration of the shape followed by a reprojection into a different grid (via trilinear interpolation), or directly via a TPS interpolator.

*Style transfer:* Let us consider two shapes $S_1$ and $S_2$, and their corresponding DVGs $V_1$, $V_2$. Projecting $S_1$ into $V_2$ performs a style transfer, in a similar fashion to AligNet[12] but without any learning — this is possible because our DVGs provide control points more adapted to the surface than AligNet's FFD grid. This use case demonstrates why we led our experiments on a chair dataset. They are strong candidates for our DVG model because:

- they generally have strong reflection symmetries;
- most details that make a chair different from another one are high resolution ones: the general aspect of a chair is not very varied (depicted in the fact that the DVG elastic hulls do not widely differ one from another).

Most of the existing work in style transfer[19,20] applies the specific (high level) details of a source shape to a target shape, while maintaining the global proportions of the target. We take an opposite view, by applying the low level information and maintaining the high level details. Our results, presented in Figure 7, show that this effectively creates meaningful objects, in two ways: the result is easy to anticipate, and corresponds to an interpretable object. New meanings for shapes can be discovered: for instance, the conformation of a chair to a sofa creates a bench (first column), which is not present in the dataset.

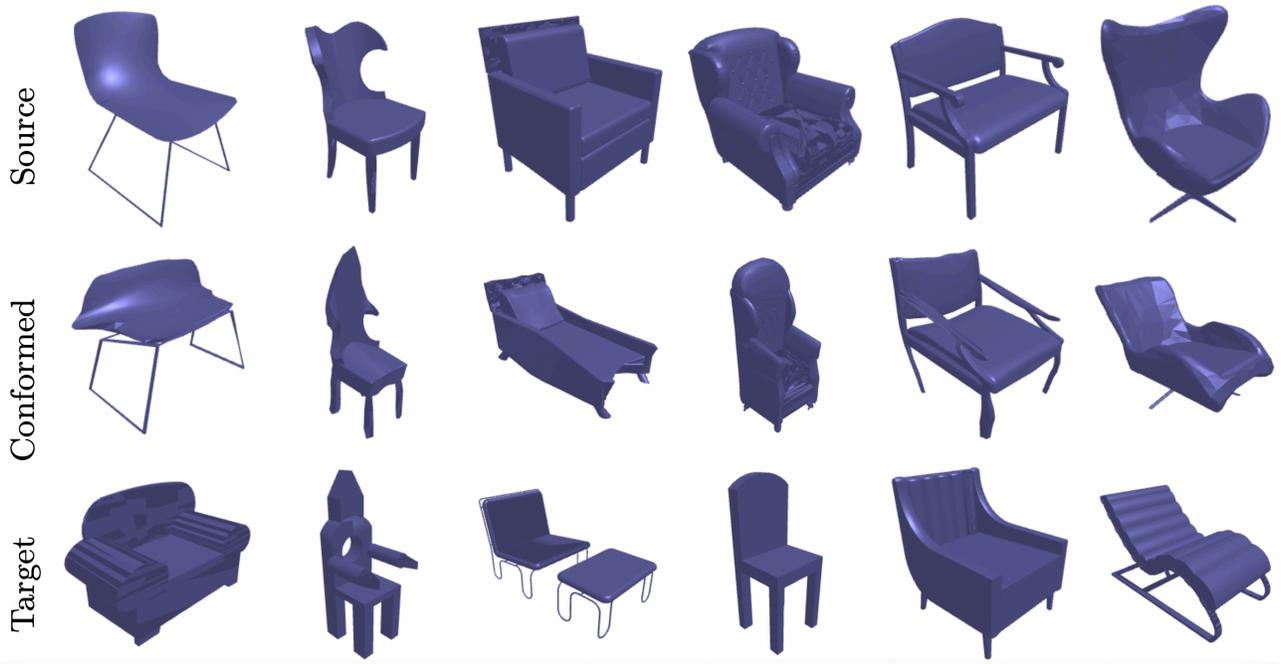

Figure 7. Style transfer performed via DVGs, via a TPS interpolator. All DVGs involved were automatically optimized using the same hyper-parameters.

*Cubification:* A special case of shape deformation is cubification — the target DVG is a straight cube. For a single-class dataset, cubification provides a consistent way of representing an object from this class; examples are shown in Figure 5c. Cubification serves further purposes, such as correspondences (see Section 3.3).

*PCA deformations:* We found that the major modes of variations for V in a dataset (the first principal components) have intuitive interpretations: for instance with chairs, the first two correspond to horizontal/vertical ratio. So, one can explore the neighborhood in PCA space of a DVG while simultaneously deforming the underlying shape. This enables an interactive shape editor where a designer can deform a shape according to intuitive parameters controlled by sliders.

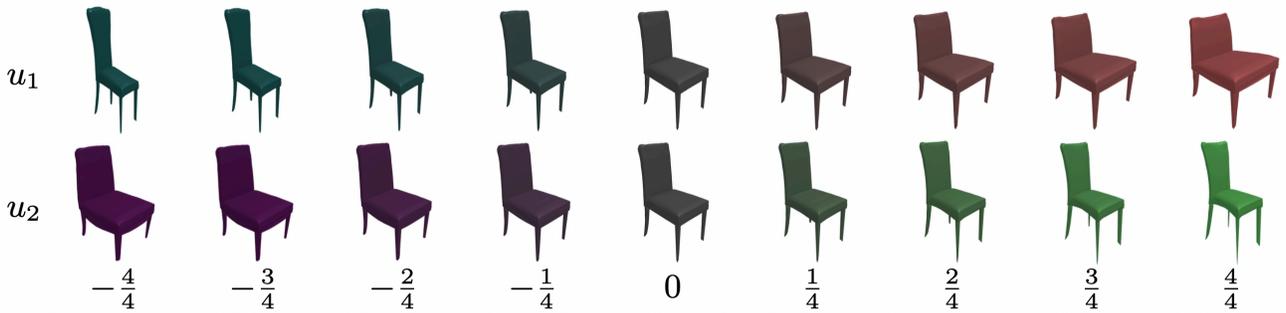

Figure 8. Having encoded k = 500 chairs with DVGs, a PCA on $\{V_k\}_k$ yields principal modes of deformations, applied to deform any shape. For example in this chair dataset, they correspond to vertical and horizontal elongations (original shape on central column; $u_i$ is the $i$-th principal component, the color gradient corresponds to the deviation from the shape's DVG).

### 3.2 Shape correspondences

Cubified shapes tend to have similar parts in similar locations, which suggests a potential for estimating shape correspondences. In order to demonstrate the versatility of our representation, we show that a naive approach already yields meaningful correspondences. In a cubified pair $(\tilde{S}, \tilde{T})$, *source* and *target*, every point $\tilde{x} \in \tilde{T}$ is matched to the closest

$\widetilde{y^\star} \in \widetilde{S}$ with respect to the Euclidean distance. Hence, for the original shapes, if we denote $C_S$ (*resp.* $C_T$) the cubification of S (*resp.* T), every $x \in T$ is matched to the solution of $\mathrm{argmin}_{y \in S} |C_T(x) - C_S(y)|_2$.

Qualitative examples of such correspondences are presented in Figure 9.

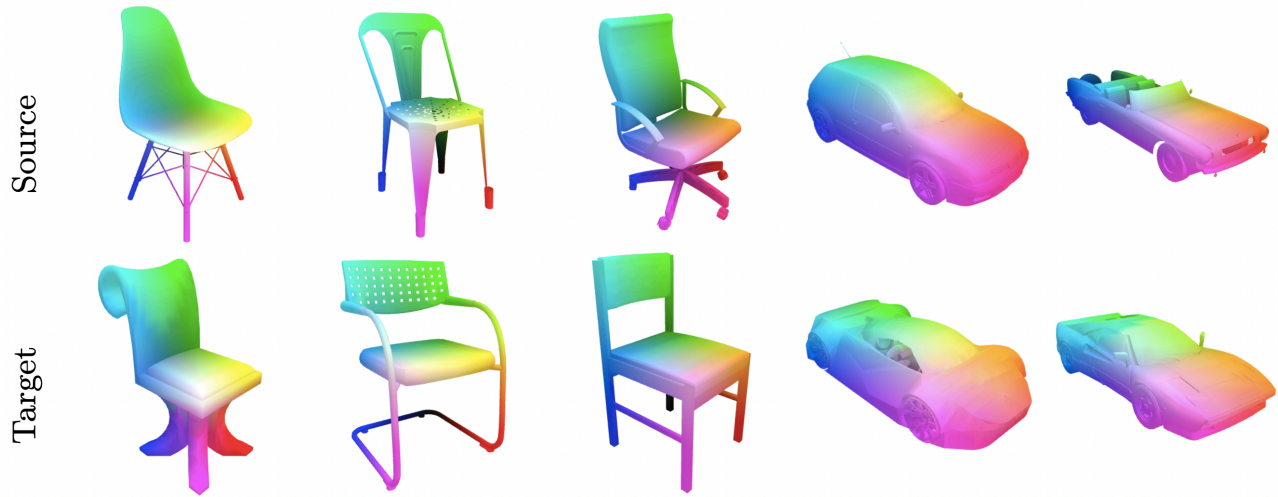

Figure 9. Shape correspondence pairs (*Source*, *Target*), estimated on cubified shapes and transferred on the original models (identified by color).

## 4. CONCLUSION

By their adaptive discretization of the embedding space, Deformable Voxel Grids offer a variety of applications for 3D shapes. They allow a natural formulation of style transfer, which effectively replaces more complex approaches relying on neural networks. While most of our results involve chairs, because of their general appearance — notably, their reflection symmetries, we would like to investigate the usefulness of DVGs for other types of shapes. For instance, we expect them to display similar results on tables, and wonder about their applications on planes (a potential hurdle being the presence of wings).

For future work, we would also like to further analyze how cubification helps improve the performance of already-existing methods. For instance, our naive correspondences could be replaced by more advanced methods (e.g. functional maps[21]), and it would be useful to test whether the correspondences obtained on a dataset can be improved with a cubification pre-processing.

Finally, cubification can also be considered as a dataset normalization step, and one could see how a classifier/segmenter neural network could benefit from it. In fact, we think that even generative networks could use this shape representation: for instance, they could learn to output a (DVG, cubified shape) pair, which would allow to condition the generation by desired style — and not only class, as is currently the case.

## ACKNOWLEDGEMENT

This work was funded in part by the French government under management of Agence Nationale de la Recherche as part of the "Investissements d'avenir" program, reference *ANR-19-P3IA-0001* (PRAIRIE 3IA Institute).